\documentclass{article}

\usepackage{booktabs}
\usepackage{a4wide}

\usepackage[hidelinks]{hyperref}

\usepackage{adjustbox}
\usepackage{graphicx}
\usepackage{amsmath, amssymb, amsthm} 
\usepackage{bm}
\usepackage{gensymb}
\usepackage{pgfplotstable}
\usepackage{tikz}
\usepackage{pgfplots}

\usepackage{diagbox}
\usepackage{multirow}
\usepackage{caption}
\usepackage{subcaption}

\usepackage{fancyhdr}
\usepackage{array}
\usepackage{wrapfig}
\usepackage{float}
\usepackage{url}

\newcolumntype{M}[1]{>{\centering\arraybackslash}m{#1}}

\newcommand{\mat}[1]{\mathbf{#1}}

\newcommand{\tens}[1]{\boldsymbol{\mathsf{#1}}}

\pgfplotsset{compat=1.18}

\begin{document}

\title{FireCastNet: Earth-as-a-Graph for Seasonal Fire Prediction}

\date{}

\author{
    Dimitrios Michail\thanks{Harokopio University of Athens, Greece. \{michail, cdavalas\}@hua.gr} \and 
    Charalampos Davalas\footnotemark[1] \and 
    Konstantinos Chafis\footnotemark[1] \and 
    Lefki-Ioanna Panagiotou\footnotemark[1] \and
    Ioannis Prapas \thanks{OrionLab, National Technical University \& National Observatory of Athens, Greece. \{iprapas, skondylatos, bountos, ipapoutsis\}@noa.gr} \and
    Spyros Kondylatos\footnotemark[2] \and 
    Nikolaos Ioannis Bountos\footnotemark[1] \footnotemark[2] \and 
    Ioannis Papoutsis \footnotemark[2]
}

\maketitle

\begin{abstract}
With climate change intensifying fire weather conditions globally, accurate seasonal wildfire forecasting has become critical for disaster preparedness and ecosystem management. We introduce FireCastNet, a novel deep learning architecture that combines 3D convolutional encoding with GraphCast-based Graph Neural Networks (GNNs) to model complex spatio-temporal dependencies for global wildfire prediction. Our approach leverages the SeasFire dataset, a comprehensive multivariate Earth system datacube containing climate, vegetation, and human-related variables, to forecast burned area patterns up to six months in advance. FireCastNet treats the Earth as an interconnected graph, enabling it to capture both local fire dynamics and long-range teleconnections that influence wildfire behavior across different spatial and temporal scales. Through comprehensive benchmarking against state-of-the-art models including GRU, Conv-GRU, Conv-LSTM, U-TAE, and TeleViT, we demonstrate that FireCastNet achieves superior performance in global burned area forecasting, with particularly strong results in fire-prone regions such as Africa, South America, and Southeast Asia. Our analysis reveals that longer input time-series significantly improve prediction robustness, while spatial context integration enhances model performance across extended forecasting horizons. Additionally, we implement local area modeling techniques that provide enhanced spatial resolution and accuracy for region-specific predictions. These findings highlight the importance of modeling Earth system interactions for long-term wildfire prediction.
\end{abstract}

\definecolor{myblue}{RGB}{0, 102, 204} 
\definecolor{mygreen}{RGB}{34, 139, 34}
\definecolor{myorange}{RGB}{255, 128, 0}
\definecolor{mycyan}{RGB}{0, 255, 255}  
\definecolor{mymagenta}{RGB}{255, 0, 255}  
\definecolor{mypurple}{RGB}{128, 0, 128} 
\definecolor{myyellow}{RGB}{200, 200, 0} 
\definecolor{mygray}{RGB}{128, 128, 128} 

\definecolor{myteal}{RGB}{0, 128, 128}
\definecolor{myseagreen}{RGB}{46, 139, 87}
\definecolor{myrose}{RGB}{188, 143, 143}
\definecolor{mydarkgreen}{RGB}{0, 100, 0}  
\definecolor{myindigo}{RGB}{75, 0, 130} 
\definecolor{mymidnight}{RGB}{25, 25, 112} 
\definecolor{mysalmon}{RGB}{250, 128, 114} 
\definecolor{mycadetblue}{RGB}{70, 130, 180}






\section{Introduction}\label{sec:intro}

Fire plays a pivotal role in the Earth system, significantly influencing ecosystems worldwide. While 
traditionally viewed as a carbon-neutral process over the long term, climate change disrupts this balance
through the intensification of fire weather conditions, leading to a rise in global fire activity~\cite{jones2022global}. 
A feedback loop is created when fires encroach into evergreen forest regions, posing a threat to their
role as carbon sinks. This situation triggers the release of stored carbon into the atmosphere, further exacerbating 
climate change~\cite{flannigan2009implications}.
Moreover, wildfires represent a critical natural hazard with far-reaching consequences for
societies worldwide, resulting in loss of life, property, infrastructure, and ecosystem
services~\cite{pettinari2020fire}.
Hence, it is imperative to improve our understanding and forecasting capabilities of wildfire phenomena within the Earth system.
By doing so, we can formulate effective strategies to mitigate the adverse impacts of wildfires and climate change. 
These strategies may encompass enhanced forestry management, strengthened infrastructure resilience, disaster preparedness,
and the implementation of more accurate warning systems.

Producing reliable long-range wildfire forecasts, spanning up to six months, requires advanced techniques that account for
the Earth as one interconnected system. This system is shaped by continuous interactions across large spatio-temporal scales.
These interactions manifest in two key ways: (a) memory effects, where past conditions like fuel accumulation and drought influence future fire activity, and (b) teleconnections, where distant climate events and weather patterns affect wildfire risk. Traditional statistical models, which rely on historical data from the same time period, have been widely used for fire hazard forecasting. However, these models often struggle to capture the intricate, non-linear patterns and dependencies that govern wildfire behavior. Given the critical importance of predicting fire occurrences to protect ecosystems and human life, it is essential to recognize and model these complex interactions effectively.

In this study, we present FireCastNet, a novel architecture aimed at enhancing seasonal wildfire prediction by utilizing the 
GraphCast~\cite{lam2023learning} weather forecasting model.  
FireCastNet aims at leveraging both short- and long-range dependencies across the Earth system and providing a deeper and more comprehensive understanding of the spatio-temporal interactions that influence wildfire behavior. Unlike traditional models, FireCastNet does not rely on proxy variables like pre-encoded oceanic indices for temporal dynamics. Instead, it directly models these dependencies from raw time series data, offering a more flexible and data-driven approach to long-range fire forecasting.

We conduct a thorough experimental evaluation, benchmarking FireCastNet's performance against baseline methods, such as the seasonal mean cycle,
and several state-of-the-art machine learning models, including Gated Recurrent Units (GRU), Convolutional GRU (Conv-GRU), Convolutional Long
Short-Term Memory (Conv-LSTM), and the U-Net Transformer-based Attention Encoder (U-TAE). This evaluation explores the predictive capabilities of
each model across various spatio-temporal contexts, assessing factors such as different time-series lengths, spatial neighborhood sizes, and
prediction horizons. Our experiments focus on forecasting burned area patterns weeks to months in advance, at a global scale and with a spatial
resolution of $0.25^{\circ}$. By leveraging spatio-temporal data at multiple scales, we assess how effectively each deep learning model captures
the intricate dynamics of wildfire behavior.

To enhance the spatial resolution and accuracy of our predictions, we employ local area modelling approaches~\cite{oskarsson2023graph, lam2023learning}. 
These techniques are integrated into FireCastNet to enable more precise localized forecasting. Our evaluation strategy involves 
a two-tier approach: first, we assess global models across regions, which are defined by the Global Fire Emissions
Database~\cite{randerson2018gfedv4} (GFED) and encompass different biomes and climatic zones. Second, we leverage local 
area modelling by training region-specific models for each GFED region, thereby further improving performance. This methodology allows us to evaluate
the model's performance across diverse geographical contexts, providing insights into its adaptability and effectiveness
in various fire-prone regions.

The main contributions of this work can be summarized as follows:
\begin{itemize}
    \item Introduction of FireCastNet, a deep learning architecture that combines 3D convolutional encoding with GraphCast-based 
          Graph Neural Networks (GNNs) to model spatio-temporal dependencies for global, seasonal wildfire prediction.
    \item Leveraging the open-access multivariate SeasFire Earth system datacube, FireCastNet forecasts global burned area patterns
          up to six months in advance, demonstrating that incorporating longer input sequences and spatial context improves predictive
          robustness compared to traditional and state-of-the-art machine learning baselines (GRU, Conv-GRU, Conv-LSTM, U-TAE, and
          TeleViT).
    \item We integrate region-specific local modelling approaches~\cite{oskarsson2023graph, lam2023learning} within FireCastNet,
          enabling higher-resolution and more accurate predictions across diverse GFED-defined biomes.
\end{itemize}
All source code can be found at \url{https://github.com/seasfire/firecastnet}.

\section{Related Work}

Wildfires are notoriously hard to model due to the non-linear interactions between the different Earth system processes
that affect them \cite{hantson2016status}. Weather, vegetation and humans interact in evolving ways that contribute in the
expansion or suppression of wildfires. Reichstein et~al.~\cite{reichstein2019deep} propose Deep Learning as a method to learn
in a  data-driven way these complex spatio-temporal interactions that influence wildfires. Several recent studies have used deep
learning for wildfire-related use cases~\cite{jain_review_2020}. For short-term daily predictions the temporal context is mostly
enough, with spatio-temporal models offering little to no advantage~\cite{prapas2021deep, kondylatos2022wildfire}. For longer-term
predictions, i.e. on subseasonal to seasonal scales, very few works have studied the effect of spatial and temporal
context~\cite{joshi2021improving, prapas2022deep, Prapas_2023_ICCV}.
Joshi et~al.~\cite{joshi2021improving} use monthly aggregated input to predict burned area using multi-layer neural networks.
Li et~al.~\cite{li2023attentionfire_v1} includes the temporal aspect of the input using a temporal attention network on time-series

Recent studies rely on the SeasFire Datacube~\cite{karasante2023seasfire} to train models with seasonal forecasting skill.
Prapas et~al.~\cite{prapas2022deep} use temporal snapshots of the fire drivers to predict future burned area patterns,
defining burned area forecasting as a segmentation task and demonstrating skillful forecasts even two months in advance. An
expansion of this~\cite{Prapas_2023_ICCV}, recognises the need to view the earth as a system for long-term forecasting and proposes a novel architecture that leverages teleconnection indices and global input in conjunction with those snapshots. This setup helps improve long-term skill, but ignores the temporal component of the input variables. 
Zhao et~al.~\cite{zhao2024causal} integrate causality with GNNs to explicitly model the causal mechanism among complex
variables via graph learning, and test their models in the European boreal and Mediterranean biome.
Finally, in the work of Zhu et~al.~\cite{zhu2025unveiling} the authors expand on the TeleViT architecture and add a balancing term to handle the data imbalance between the burned and non-burned classes. This allows them to create a fire risk index that is better calibrated between different regions. 

Machine learning has recently made significant advances in global weather forecasting~\cite{rasp2024weatherbench}, with graph
neural networks showing tremendous potential to represent the Earth as a system~\cite{keisler2022forecasting, lam2022graphcast}.
Keisler~\cite{keisler2022forecasting} introduces a GNN-based model that predicts multiple atmospheric variables across
various pressure levels on a global scale. 
Building upon this foundation, Lam et~al.~\cite{lam2022graphcast} introduced the GraphCast model, a machine
learning method that forecasts
hundreds of weather variables over a 10-day period at 0.25-degree resolution globally, delivering results in under a minute.
Further advancing the GraphCast architecture, the work of Oskarsson et~al.~\cite{oskarsson2024probabilistic} introduces Graph-EFM, a probabilistic
weather forecasting model that combines a flexible latent-variable formulation with a hierarchical graph-based framework. This
model efficiently generates spatially coherent ensemble forecasts, achieving errors equivalent to or lower than comparable
deterministic models while accurately capturing forecast uncertainty. These types of models  have also been shown to be
effective at the temporal scales that we are interested in this study, with Fuxi-S2S~\cite{chen2023fuxi} achieving skillfull
forecasts at the subseasonal to seasonal scales. Our work aims to leverage the advances in data-driven weather modeling for
burned area forecasting with an architecture inspired by Graphcast.

Complementing global approaches, the paradigm of Local Area Modeling has emerged as a powerful strategy for improving forecast
skill in limited geographic regions. Oskarsson et~al.~\cite{oskarsson2023graph} propose a graph-based neural weather prediction
method that tailors modeling to limited areas rather than the entire globe. This approach focuses computational resources locally,
enabling higher resolution and potentially finer-grained feature learning in regions of interest, which is particularly
beneficial when computational constraints or application domain focus (e.g., wildfire-prone areas) restrict the scope of
forecasts.

\section{Seasonal Fire Forecasting}

\subsection*{Problem Formulation}

We view the problem as binary classification at a particular location of the cube
and a particular timestamp. Thus, given a triplet $(\phi_c, \lambda_c, t_c)$ of latitude, longitude and $8$-day period we
predict whether a fire will occur or not at that particular location in time and space. As input we use a timeseries for
each variable of length $ts \in \{6, 12, 24\}$ timesteps, where each timestep corresponds to a single $8$-day period.
As target variable  
we predict the presence of burned areas at a future timestep $t+h$ with different 
values of $h \in \{1, 2, 4, 8, 16, 24\}$.

\subsection*{Dataset}\label{sec:dataset}

\begin{table}[t]
\centering
\begin{tabular}{llm{6em}}
\toprule
\textbf{Input type} &  \textbf{Variable} & \textbf{Abbreviation}\\
\midrule
\multirow{10}{2.5em}{\textbf{Basic Input}} & Mean sea level pressure & mslp \\
 &Total precipitation & tp \\
 &Vapor Pressure Deficit & vpd \\
 &Sea surface temperature & sst \\
 &Temperature at 2 meters - Mean & t2m\_mean \\
 &Surface solar radiation downwards & ssrd \\
 &Volumetric soil water level 1 & swvl1 \\
 &Land surface temperature at day & lst\_day \\
 &Normalized Difference Vegetation Index & ndvi \\
 &Population density & pop\_dens \\
 \midrule
 \textbf{Static input} & Land Sea Mask & lsm \\
  \midrule
 \multirow{2}{3em}{\textbf{Positional input}} &Latitude (sin/cos) &  lat \\
 &Longitude (sin/cos) & lon  \\
 \midrule
 \textbf{Target} & Burned Areas (as binary) & gwis\_ba\\
 \bottomrule
\end{tabular}

\caption{Fire driver variables (10 input, 1 static and 2 positional) used for predictions.
Target variable represents burned area in hectares (converted to binary for classification tasks).}
\label{tab:variables}
\end{table}

The SeasFire Datacube~\cite{karasante2023seasfire} is an analysis-ready, open-access datacube fit for wildfire forecasting at different spatio-temporal scales that serves not only for seasonal fire forecasting worldwide but also for forecasting emissions from wildfires and predicting the evolution of wildfire regimes.
It spans over 21 years (2001-2021) with an 8-day temporal and a $0.25^{\circ}$ spatial resolution. 
The dataset provides a comprehensive coverage of atmospheric, climatological, vegetation and socioeconomic factors influencing wildfires and contains
58 variables in total, along with target variables,
such as burned areas, fire radiative power, and carbon emissions from wildfires. 
11 different variables were utilized (see Table~\ref{tab:variables}) in order to predict 
the existence of fire in different time horizons. In addition and when needed, we also used simple positional encodings by
augmenting the feature vector with the actual cube coordinates, i.e. sin/cos of latitude and longitude. 
All variables, except the cube coordinates, were standardized before use. 

\subsection*{Naive Forecasting}\label{sec:naive}

When dealing with seasonal time 
series, the naive seasonal forecasting method uses the observed values in the 
same period of the previous seasons~\cite{hyndman2018forecasting}. For example 
in order to predict for a particular week of August in a particular year, one could use the 
observed value in the same week of August one year before. Our dataset contains multiple years, allowing for different
baseline methods based on the utilization of past data. We utilize the following two: 
\begin{itemize} 
   \item To forecast the occurrence of fire at a particular location in a particular 8-day
   period of the year, we examine whether there were fires at that location in the specific 8-day period in any of the
   previous years. If there were, the model predicts a fire; otherwise, it predicts no fire. We call this baseline the 
   {\em any-fire} baseline.
   \item To forecast the occurrence of fire at a particular location in a particular 8-day
   period of the year we use a majority rule. Thus,  
   we predict fire if the number of previous years with fire in that particular 8-day period is larger than the number of
   previous years without fire at that period. This baseline is called the {\bf majority} baseline.   
\end{itemize}
The majority baseline approach aligns effectively with the binary classification task.

\subsection{Baseline Models}

\subsubsection*{Gated Recurrent Units}

Recurrent Neural Networks~\cite{medsker2001recurrent} are a natural choice when trying to capture temporal dependencies.
We utilize a very simple architecture comprising of Gated Recurrent Units (GRU). The neural network is
comprised of two GRU layers with 64 hidden channels in each layer. A dropout layer with a probability of 0.1 exists between
the two GRU layers. As the task is binary classification, a final linear layer reduces the representation into a single
prediction and a sigmoid function outputs the final prediction.

\subsubsection*{Convolutional GRU}\label{sec:conv-gru}

To capture both temporal and spatial dependencies at the same time, we utilize a convolutional 
GRU network~\cite{DBLP:journals/corr/BallasYPC15} (Conv-GRU).
In such a network all inputs, cell outputs, hidden states 
and gates are 3D tensors where the last two dimensions are spatial dimensions (rows and columns).
This architecture is particularly effective for spatio-temporal tasks, as it incorporates convolution operations
in the update and reset gates to model spatial relationships.

The Conv-GRU model is defined using the following set of equations, where $*$ denotes the convolution operator,
\( \odot \) is element-wise multiplication, and \( \sigma \) is the sigmoid activation function:
\[
\begin{aligned}
    \tens{z}_t & = \sigma(\mat{W}_{xz} * \tens{x}_t + \mat{W}_{hz} * \tens{h}_{t-1} + \tens{b}_z), \\
    \tens{r}_t & = \sigma(\mat{W}_{xr} * \tens{x}_t + \mat{W}_{hr} * \tens{h}_{t-1} + \tens{b}_r), \\
    \tilde{\tens{h}}_t & = \tanh(\mat{W}_{x\tilde{h}} * \tens{x}_t + \tens{r}_t \odot (\mat{W}_{h\tilde{h}} * \tens{h}_{t-1}) + \tens{b}_{\tilde{h}}), \\
    \tens{h}_t & = (1 - \tens{z}_t) \odot \tens{h}_{t-1} + \tens{z}_t \odot \tilde{\tens{h}}_t.
\end{aligned}
\]
Here, \( \tens{x}_t \in \mathbb{R}^{F \times H \times W} \) represents the input tensor at time step \( t \), 
where \( F \) is the number of features, and \( H \) and \( W \) are the spatial dimensions. 
\( \tens{h}_{t-1} \) is the hidden state from the previous time step, while \( \tens{z}_t \) and \( \tens{r}_t \) are the update
and reset gates, respectively. The hidden state \( \tens{h}_t \) is updated by controlling how much of the past hidden
state is carried forward (via \( \tens{z}_t \)) and how much of the new information (via \( \tilde{\tens{h}}_t \)) is incorporated
into the current state.

Given $(\phi_c, \lambda_c)$ for a particular location of interest and a particular time $t$ we generate
a 3D tensor $\tens{x}_t$ whose 
last two dimensions are the number of rows and columns of a $(2r+1) \times (2r+1)$ spatial grid. The 
first dimension is the number of features.
The grid is centered at our location of interest
while the remaining locations represent $(\phi_c \pm i \cdot 0.25^{\circ}, \lambda_c \pm i \cdot 0.25^{\circ})$
for $i \in \{1, \ldots, r-1\}$. We feed $\tens{x}_t$ to the model for each timestep $t$ in order to compute 
the hidden representation $\tens{h}_t$.
The final result is obtained by utilizing multilayer perceptron (MLP) layers in order to gradually aggregate
the information into a single prediction.

\subsubsection*{Convolutional LSTM}\label{sec:conv-lstm}

We also utilize a convolutional LSTM network~\cite{shi2015convolutional} (Conv-LSTM). Again all inputs,
cell outputs, hidden states 
and gates are 3D tensors where the last two dimensions are spatial dimensions (rows and columns).
The future state of a certain cell is computed from the input and the past states of its local 
neighbors. This is achieved by using convolution operators in different transitions such as the
state-to-state and/or input-to-state. 

We utilize a slightly different network from the one in Shi et~al.\cite{shi2015convolutional}, which is implemented using the following equations:
\begin{align*}
   \tens{i}_t & = \sigma(\mat{W}_{xi} * \tens{x}_t + \mat{W}_{hi} * \tens{h}_{t-1} + \tens{b}_i)\\
   \tens{f}_t & = \sigma(\mat{W}_{xf} * \tens{x}_t + \mat{W}_{hf} * \tens{h}_{t-1} + \tens{b}_f)\\
   \tens{o}_t & = \sigma(\mat{W}_{xo} * \tens{x}_t + \mat{W}_{ho} * \tens{h}_{t-1} + \tens{b}_o)\\
   \tens{g}_t & = \tanh(\mat{W}_{xg} * \tens{x}_t + \mat{W}_{hg} * \tens{h}_{t-1} + \tens{b}_g)\\
   \tens{c}_t & = \tens{f}_t \odot \tens{c}_{t-1} + \tens{i}_t \odot \tens{g}_t\\
   \tens{h}_t & = \tens{o}_t \odot \tanh(\tens{c}_t)
\end{align*}
Here $*$ denotes the convolution operator and $\odot$ point-wise multiplication.
We integrate the Conv-LSTM into the classification pipeline the same way as the Conv-GRU architecture.

\subsubsection*{U-Net with Temporal Attention Encoder}\label{sec:utae}

The U-Net with Temporal Attention Encoder (U-TAE)~\cite{garnot2021panoptic} architecture is designed for spatio-temporal
feature extraction
and segmentation of satellite image time series. It extends the classic U-Net by incorporating a temporal
self-attention mechanism to process temporal sequences of images. This enables the network to capture temporal
dependencies across multiple resolution levels.

The U-TAE consists of two main components: the spatial encoder-decoder network and the temporal self-attention module.
The spatial encoder-decoder network follows the U-Net architecture and processes each image in the time series separately.
The encoder applies convolutional layers to extract spatial features from the images at different resolutions. Specifically,
each encoder block consists of two $3 \times 3$ convolutions followed by Group Normalization and ReLU activations, halving
the spatial resolution at each level. The decoder performs upsampling using transposed convolutions, and the corresponding
encoder feature maps are concatenated at each level (U-Net-style skip connections).
To capture the temporal dynamics, U-TAE incorporates a Lightweight-Temporal Attention Encoder~\cite{garnot2020lightweight} (L-TAE)
module, which is applied independently to each pixel's feature map sequence extracted from the lowest-resolution level of the
spatial encoder. The L-TAE computes temporal attention masks over the sequence, allowing the model to focus on relevant time
steps. These temporal attention maps are then bilinearly interpolated to match the resolutions of the higher levels of the
encoder. After the attention is applied to each feature map group, the feature maps are concatenated and passed through
a $1 \times 1$ convolution to fuse temporal information.

\section{FireCastNet}

\begin{figure*}[t]
    \centering
    \includegraphics[width=\columnwidth]{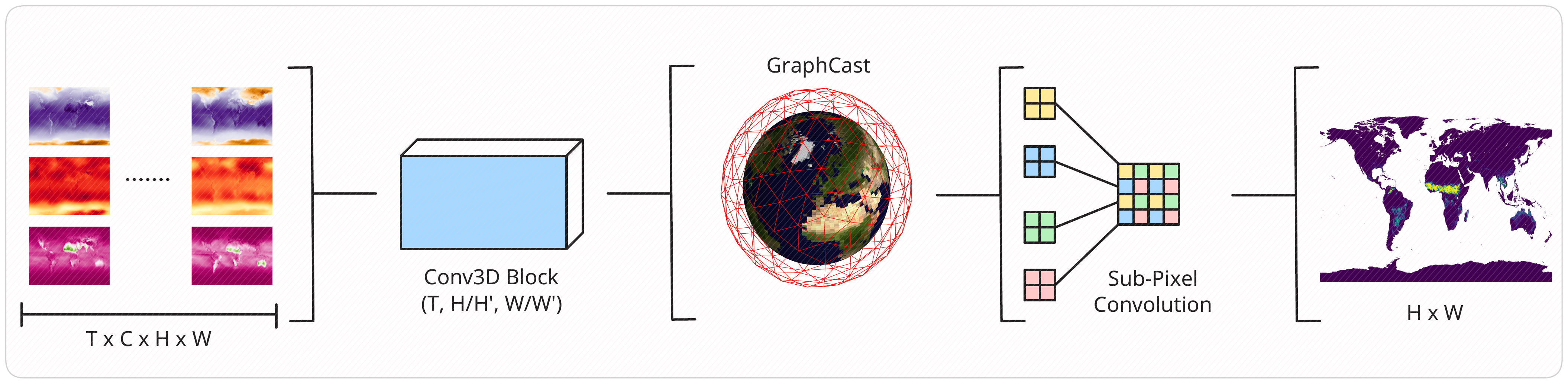} 
    \caption{High-level representation of our architecture.
The FireCastNet architecture consists of three main components: (a) a cube embedding block for spatio-temporal feature extraction using
a 3D convolution layer that reduces the spatial and temporal dimensions of the input timeseries data 
(b) a GraphCast block, which leverages a multi-mesh Graph Neural Network (GNN) to model long-range spatial interactions across a
refined icosahedral mesh structure, and (c) a sub-pixel convolution block for upscaling the output to match the resolution of the
original input.
}
    \label{fig:high_level_representation}
\end{figure*}

\begin{figure*}[htbp]
    \begin{subfigure}[t]{0.49\textwidth}
        \centering
        \includegraphics[width=\textwidth]{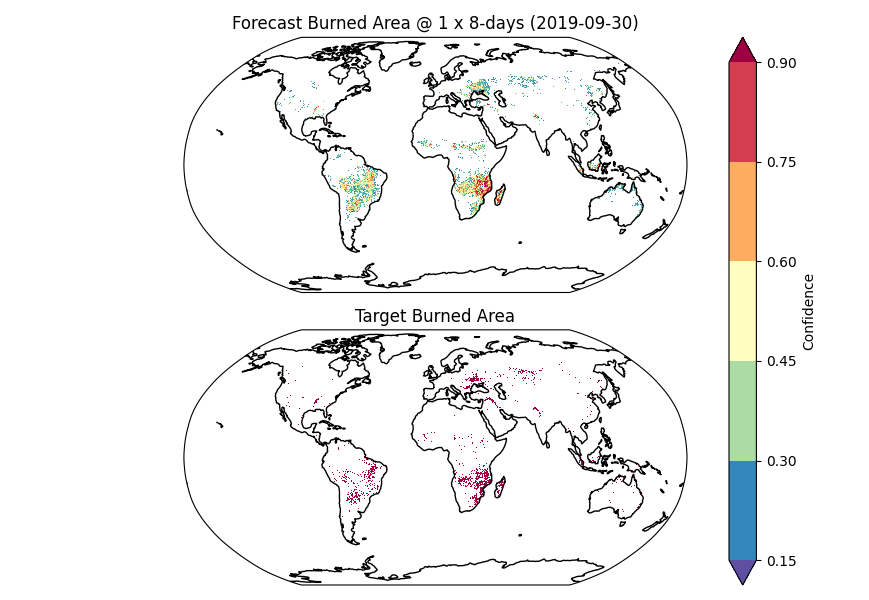} 
    \end{subfigure}%
    \hfill
    \begin{subfigure}[t]{0.49\textwidth}
        \centering
        \includegraphics[width=\textwidth]{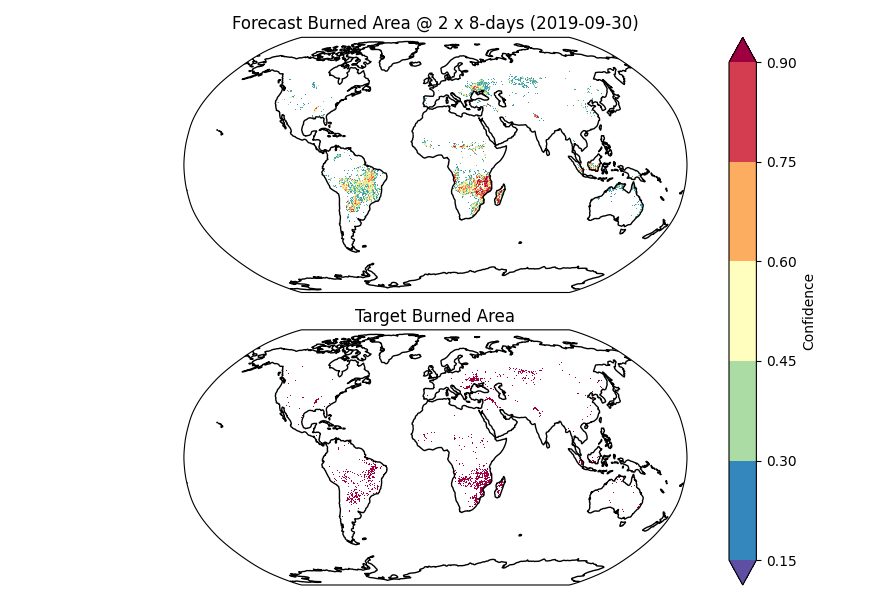}         
    \end{subfigure}%
    \vspace{3pt}
    \begin{subfigure}[t]{0.49\textwidth}
        \centering
        \includegraphics[width=\textwidth]{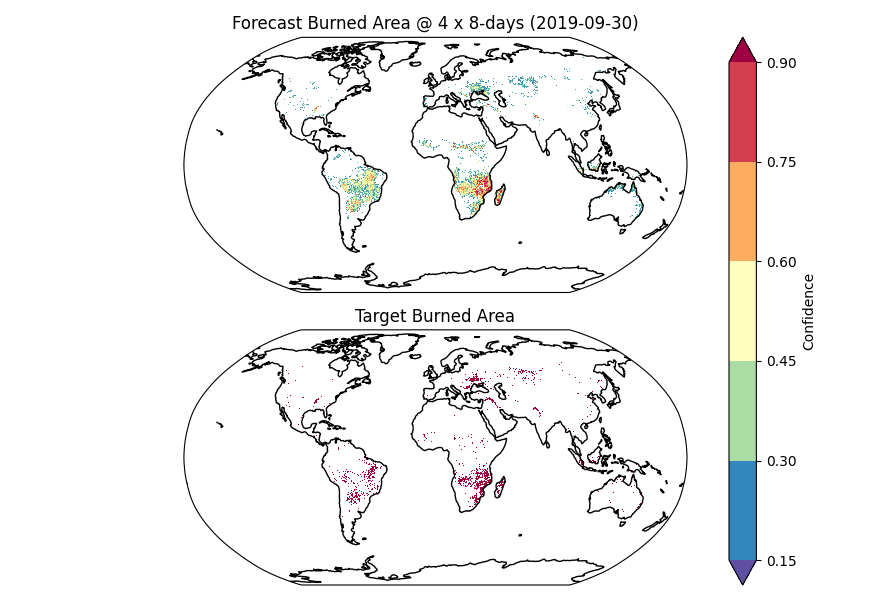}                 
    \end{subfigure}%
    \hfill
    \begin{subfigure}[t]{0.49\textwidth}
        \centering
        \includegraphics[width=\textwidth]{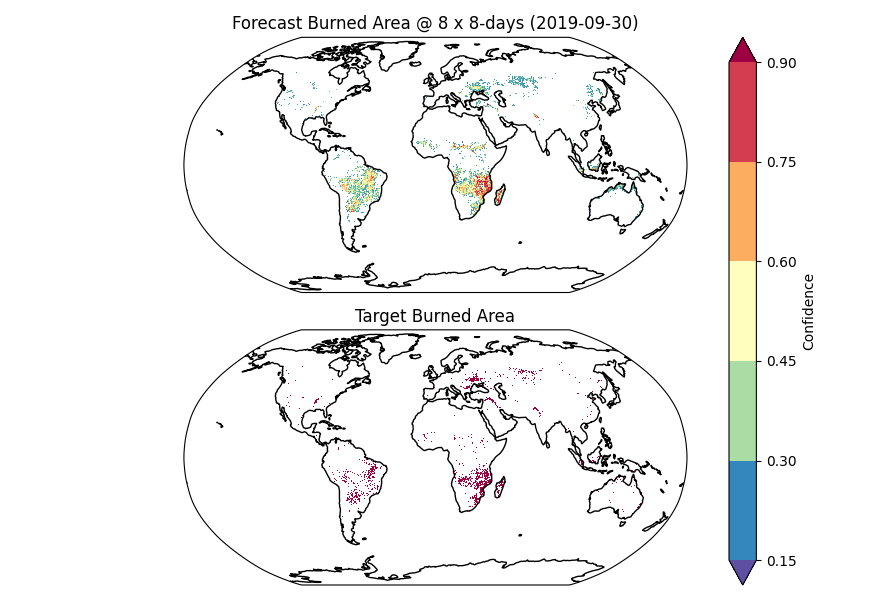}                         
    \end{subfigure}%
    \vspace{3pt}
    \begin{subfigure}[t]{0.49\textwidth}
    \centering
        \includegraphics[width=\columnwidth]{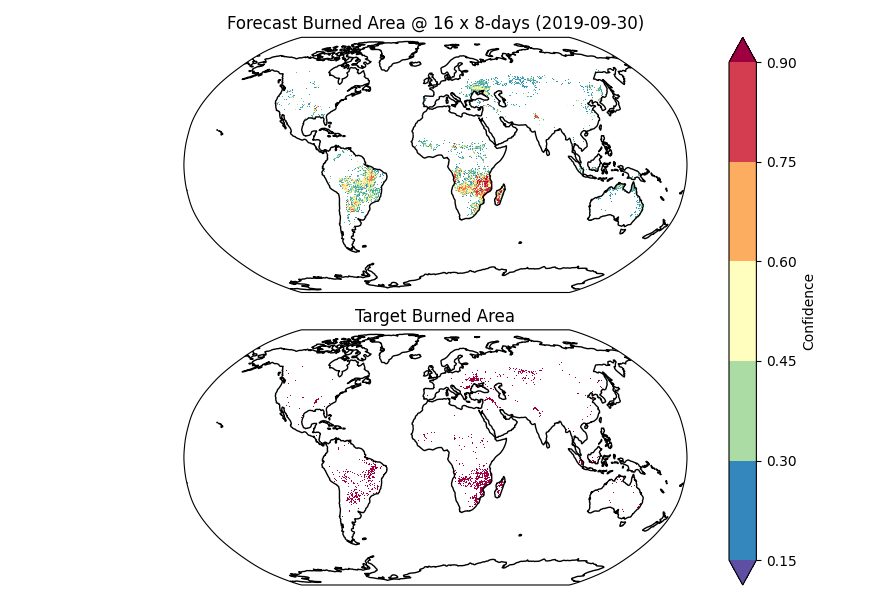}                             
    \end{subfigure}%
    \hfill
    \begin{subfigure}[t]{0.49\textwidth}
    \centering
        \includegraphics[width=\columnwidth]{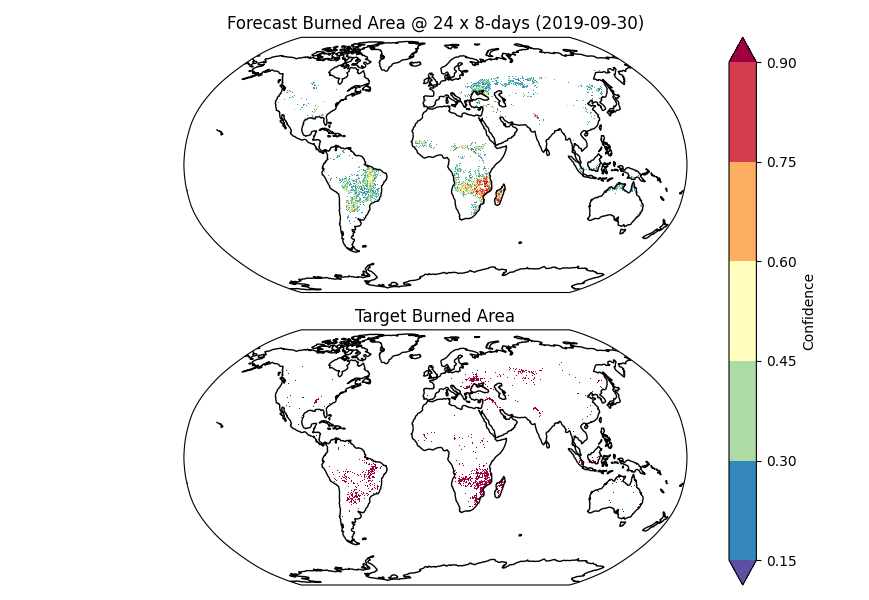}                             
    \end{subfigure}
    \caption{
        Target variable and prediction (per horizon) of the best global model, using a time-series of length 24,
        on 30 September 2019. The target variable is always the same. The input variables depend on the prediction 
        horizon, i.e. for $h=1$ we use a time-series of length 24 ending one $8$-day period before, for $h=2$ we use
        a time-series of length 24 ending two $8$-day periods before, and so on.
        The sigmoid output represents the prediction confidence.
    }
    \label{fig:firecastnet:example:prediction}
\end{figure*}

 The FireCastNet architecture is shown in Figure~\ref{fig:high_level_representation}. It consists of a
 cube embedding block, a GraphCast block and a sub-pixel convolution for upsampling.
  This architecture is designed to
 perform segmentation tasks on a timeseries of multi-channel images. The input data is represented as a tensor of dimensions
 \( \tens{x} \in \mathbb{R}^{T \times C \times H \times W} \), where \( T \) is the number of temporal steps, \( C \) is the number of input
 channels, and \( H \) and \( W \) are the height and width of each image, respectively.
 For reference, ERA5 has spatial dimensions \(721 \times 1440\), while the SeasFire datacube uses \(720 \times 1440\). Thus, in our case, 
 we have $H = 720$ and $W = 1440$.

\subsection{Cube Embedding}

The first stage of FireCastNet is a 3D convolutional block designed for spatio-temporal feature extraction. Its space–time cube-embedding
reduces the spatial and temporal dimensions of the input, thereby accelerating training~\cite{tong2022videomae}.
This embedding applies a 3D convolution with kernel and stride $T \times \frac{H}{H'} \times \frac{W}{W'}$, 
and increases the channel dimension from $C$ to $C' > C$.
Formally, the block is expressed as $\tens{x}' = \text{Conv3D}(\tens{x}; \theta)$, and produces an output tensor 
\( \tens{x'} \in \mathbb{R}^{C' \times H' \times W'} \). 

Maintaining a spatial resolution of \(0.25^\circ\) requires \(\frac{H}{H'} = 1\) and \(\frac{W}{W'} = 1\). Instead, we reduce the
resolution to \(1^\circ\) by setting \(\frac{H}{H'} = 4\) and \(\frac{W}{W'} = 4\). The model uses 11 variables from the datacube
(Table~\ref{tab:variables}), along with additional channels containing \(\cos(\text{latitude})\) and \(\sin/\cos(\text{longitude})\),
giving \(C = 14\). We expand this to \(C' = 64\). Since $H'=180$ and $W'=360$, the resulting cube-embedding output is
\( \tens{x'} \in \mathbb{R}^{64 \times 180 \times 360} \). 

\subsection{GraphCast}

The core of the FireCastNet architecture is GraphCast, which uses a Graph Neural Network (GNN) in an "encode-process-decode"
configuration. To capitalize on the GNN's ability to handle arbitrary sparse interactions, it uses an internal "multi-mesh" graph representation.
This structure facilitates the modeling of long-range interactions within a few message-passing steps and provides a more homogeneous spatial resolution
across the globe compared to a standard latitude-longitude grid. The latitude-longitude grid is a less suitable representation due to its spatial inhomogeneity
and the high density of grid points at the poles, which demands disproportionate computational resources.

The multi-mesh is a spatially homogeneous graph with high spatial resolution over the globe. It is constructed by iteratively refining a regular
icosahedron (a polyhedron with 12 nodes, 20 faces, and 30 edges) six times. In each refinement step, every triangular face is subdivided into four
smaller triangles by connecting the midpoints of its edges, leading to four times more faces and edges. The newly created vertices are then
reprojected onto a circumscribing sphere to ensure the mesh closely approximates the shape of the Earth. This hierarchical process yields a series
of increasingly finer meshes. The multi-mesh ultimately contains the 40,962 nodes from the highest resolution mesh, and the union of all the edges
created in the intermediate graphs, forming a flat hierarchy of edges with varying lengths. This multi-scale graph allows coarse edges to bridge
long distances and fine edges to capture local interactions.

The module operates in three stages:
\begin{itemize}
    \item \textbf{Encoder:} The encoder maps the input data from the latitude-longitude grid to the multi-mesh. It uses a GNN with directed edges from the
    grid points to the multi-mesh nodes to encode the grid-based features into learned node attributes. At this stage, additional positional information for
    the mesh nodes and edges is also encoded along with the features from the grid. The output is an initial set of node embeddings on the multi-mesh,
    \( \tens{h}_0 \in \mathbb{R}^{|V| \times C_{\text{mesh}}} \), where \(|V|\) is the number of mesh nodes. We use $C_{\text{mesh}} = 64$.

    \item \textbf{Processor:} The processor consists of 12 consecutive GNN layers that perform learned message-passing on the multi-mesh graph. Each layer
    updates the node embeddings, allowing information to propagate efficiently across space. The long-range edges in the multi-scale mesh structure are
    crucial for capturing distant teleconnections. The update at each layer \(l\) can be represented as
    $\tens{h}_{l+1} = \text{GNN}_{\text{process}}(\tens{h}_l, G)$,
    where \(G\) is the multi-mesh graph.

    \item \textbf{Decoder:} Finally, the decoder maps the processed node embeddings from the multi-mesh back to the latitude-longitude grid. It employs a
    GNN with directed edges from the multi-mesh nodes to the grid points. This grid representation is then combined with the input state to form the final
    output prediction. The decoding step is represented as $\tens{x}_{\text{dec}} = \text{GNN}_{\text{dec}}(\tens{h}_{12}; G)$.
\end{itemize}

\subsection{Upscaling}

The final stage of FireCastNet is a sub-pixel convolution~\cite{shi2016real} block, which upscales the decoded
output \( \tens{x}_{\text{dec}} \) to match the spatial dimensions of the original input. 
We choose the number of channels of the decoder to be \( r^2 \),
where \( r \) is the upscaling factor (in our case, \( r = \frac{H}{H'} = \frac{W}{W'} = 4 \) to restore the original \(0.25^\circ\) resolution).
Thus, the decoder produces \( \tens{x}_{\text{dec}} \in \mathbb{R}^{16 \times 180 \times 360} \). The sub-pixel convolution 
rearranges the feature maps into higher-resolution output, producing the final segmentation map
of size \( H \times W  = 720 \times 1440 \). The upscaling operation is defined as
$\tens{x}_{\text{out}} = \text{Sub-Pixel-Conv}(\tens{x}_{\text{dec}})$,
yielding the segmentation output.

\begin{table*}[t]

\begin{tabular}{|l|c|c|c|c|c|c|}
    \toprule
    \textbf{Model \; \textbackslash \; Horizon} & 1 & 2 & 4 & 8 & 16 & 24 \\
    \midrule
    GRU                & 0.558        & 0.549        & 0.540        & 0.519    & 0.521     & 0.520 \\
    Conv-GRU           & 0.593        & 0.579        & 0.556        & 0.551    & 0.545    & 0.569      \\
    Conv-LSTM          & 0.635  & 0.608        & 0.615        & 0.601    & 0.598     & 0.589 \\
    U-TAE               & 0.624        & 0.621        & 0.611        & 0.601    & 0.608     & 0.594      \\
    TeleViT~\cite{Prapas_2023_ICCV} & 0.622        & 0.617        & 0.610        & 0.611   & 0.606     & 0.604 \\
    FireCastNet (ours) & {\bf 0.641}        & {\bf 0.636}  & {\bf 0.631}  & {\bf 0.631} & {\bf 0.628} & {\bf 0.633} \\
    \bottomrule
\end{tabular}
\centering
\caption{Comparison of all global-scale models. The time-series length is 24 (except for TeleViT, which
    cannot process time-series). }  \label{tab:all:ts24:globe:auprc}
\end{table*}

\section{Experiments}\label{sec:experiments}
   
Our methodology can be directly applied to any region of the world, producing different models per region. For our forecasting task,
our split is time-based, using years 2002-2017, 2018 and 2019 for train, validation, and test, respectively.
The SeasFire dataset is utilized in all the experiments (see Section~\ref{sec:dataset}).
Figure~\ref{fig:firecastnet:example:prediction} shows the ground truth and model predictions for September 30, 2019. 
For each time horizon $h=\{1, 2, 4, 8, 16, 24\}$, we train a separately model. 
Each such model uses a time-series of length $ts=24$ ending $h$ periods before the target date, predicting
the presence of fire in the next $h$ 8-day period.

Due to our highly imbalanced dataset, the performance of the model is evaluated using the Area Under the Precision-Recall Curve
(AUPRC)~\cite{davis2006relationship}. We calculate the metric using the average precision score, which summarizes a precision-recall
curve as a weighted mean of precisions at each threshold, with the difference in recall from the previous threshold as weight
\[
  AP = \underset{t}{\sum} (R_t - R_{t-1}) \cdot P_t, 
\]
where $P_t, R_t$ is the respective precision and recall at threshold index $t$.

\begin{figure*}[t]
    \centering
    \includegraphics[width=0.8\columnwidth]{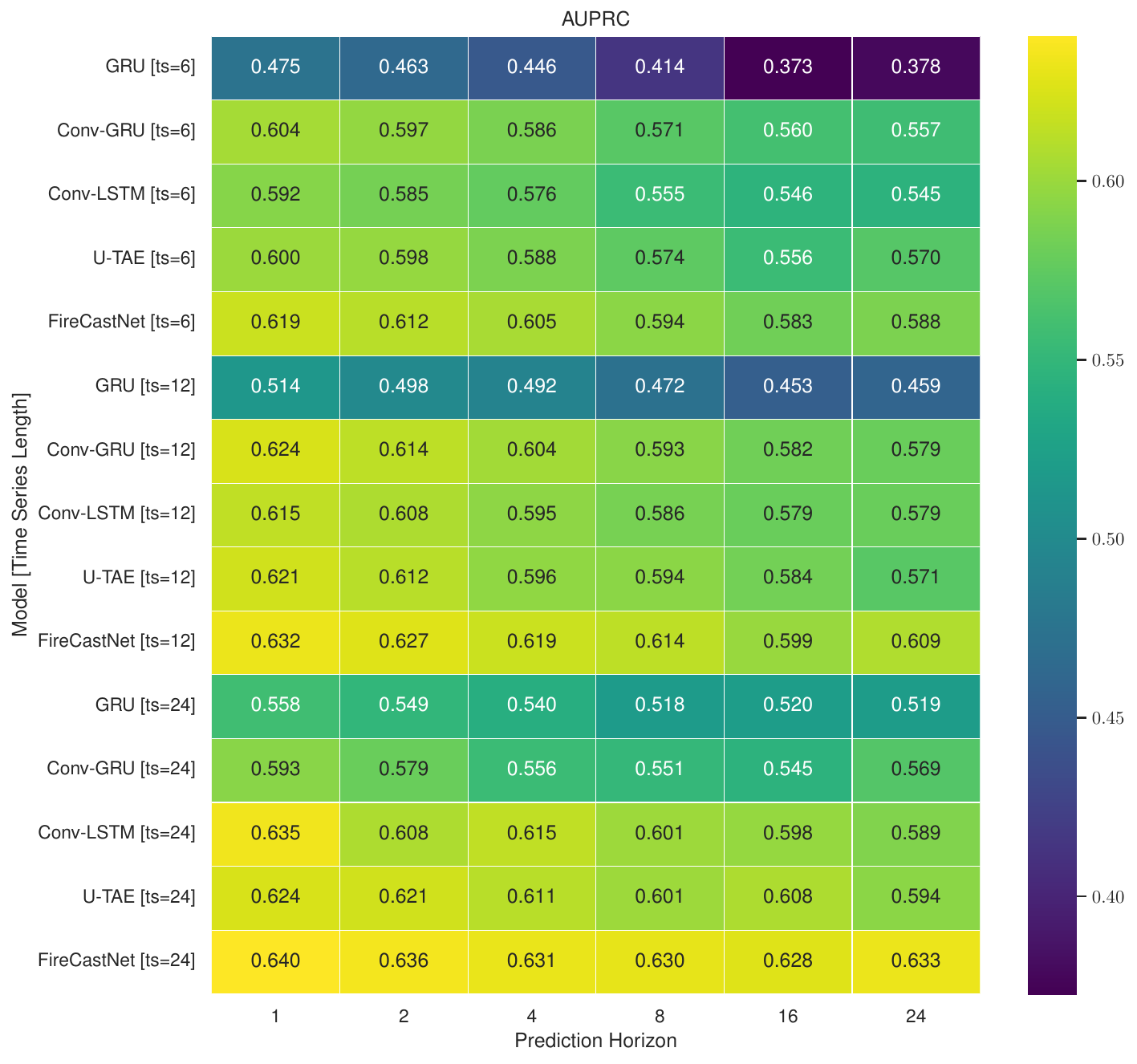} 
    \caption{Model performance at a global scale with different time-series length $ts \in \{6, 12, 24\}$ periods (8-days).}\label{fig:global:timeseries}
    \label{fig:auprc:heatmap:all:models:global}
\end{figure*}

Models are trained for 50 epochs. We use binary cross-entropy loss and the
AdamW~\cite{DBLP:conf/iclr/LoshchilovH19} optimizer. The initial learning rate is set to $0.001$ and adjusted using
SGDR~\cite{DBLP:conf/iclr/LoshchilovH17}. The schedule contains two cycles with 10 and 40 epochs, respectively. A weight decay
factor of $10^{-7}$ is used when training all models.

\subsection{Global Model}

In this section, we train and compare FireCastNet at a global scale. 
The naive prediction baseline models discussed in Section~\ref{sec:naive} 
exhibit $0.174$ (anyfire) and $0.3211$ (majority) AUPRC. The first baseline checks whether any previous year
contains some fire in the same period, while the second checks whether a majority of the previous years
contains fires. 

All models take as input the 11 inputs described in Table~\ref{tab:variables}. 
The GRU model is comprised of a single hidden layer with 64 hidden units.   
The Conv-GRU and Conv-LSTM models use the same parameters as the GRU with the only difference of an additional
setting of the $5 \times 5$ kernel. The U-TAE model is configured with 16 attention heads, and a layer size of 256.
Each attention head has a key dimension of 4.
      
Finally, FireCastNet processes data with spatial resolution $0.25^{\circ}$ and geographic bounds from -89.875° to
89.875° latitude and -179.875° to 179.875° longitude. The Conv3d block decreases spatial resolution to $1^{\circ}$,
and adjusts the hidden dimensionality to 64, with layer normalization.  
Additional mesh input includes 3 channels for nodes and 4 for edges. The model's processor features 12 layers,
and a hidden layer dimension of 64.

Table~\ref{tab:all:ts24:globe:auprc} provides results between all baselines and the FireCastNet
model for timeseries length 24 for different time horizons. The TeleVit~\cite{Prapas_2023_ICCV} model is also
included in the comparison. TeleViT is a teleconnection-driven vision transformer for seasonal wildfire forecasting,
trained also on the SeasFire dataset. It should be noted that it does not work directly with timeseries data. Instead,
it extracts temporal context from Oceanic Indices (OCIs). In all cases, different models were trained for each 
prediction time horizon.
Regardless of the range of forecasts, the FireCastNet is clearly more potent in per-pixel classification.

\begin{figure}[t]
        \centering
        \begin{tikzpicture}
            \begin{axis}[
                xlabel={Forecasting Window Length (8-days period)},
                ylabel={AUPRC (\%)},
                legend style={at={(0.42,0.01)}, anchor=south, legend columns=5},
                xtick={1,2,4,8,16,24},
                ytick={0.4, 0.45,0.5, 0.55,0.6, 0.65,0.7, 0.75, 0.8, 0.85,0.9,1.0},
                xmin=0,
                xmax=25,
                ymin=0.50, 
                ymax=0.65,
                grid=both,
                major grid style={line width=.1pt,draw=mygray!50},
                minor tick num=1,
                every major grid/.style={line width=.1pt,draw=mygray!50},
            ]
            \addplot[style={solid}, mark=diamond, color=red] table [x=x, y=ts24overlap3] {firecastnet-globe.dat};
            \addplot[mark=pentagon, color=mypurple] table [x=x, y=ts24overlap6] {firecastnet-globe.dat};
            \addplot[mark=triangle, color=blue] table [x=x, y=ts24overlap12] {firecastnet-globe.dat};
            \addplot[mark=star, color=myorange] table [x=x, y=ts24overlap18] {firecastnet-globe.dat};
            \addplot[mark=square, color=mygreen] table [x=x, y=ts24] {firecastnet-globe.dat};
            \legend{3, 6, 12, 18, 23}

            \end{axis}
        \end{tikzpicture}
        \caption{Performance of FireCastNet model at
        a global scale with timeseries length $24$ for 
        different time overlap windows.}\label{fig:all:ts24:globe:auprc:overlap}
    \end{figure}

\subsection{Timeseries Length}

In this section, we investigate the impact of training global models using input time series of varying lengths, both with and without spatially
accumulated information. This analysis aims to explore how the temporal context provided to the models affects their predictive performance across
different forecasting horizons. Specifically, we evaluate models trained with three distinct time-series lengths, $ts \in \{6, 12, 24\}$, and
analyze their performance in terms of the Area Under the Precision-Recall Curve (AUPRC). For each combination of time-series length and forecasting
horizon $h \in \{1, 2, 4, 8, 16, 24\}$, we train a separate model to assess the influence of these parameters independently.

Figure~\ref{fig:global:timeseries} summarizes the performance of the models under these configurations, offering insights into how time-series
length affects prediction accuracy. Among the various models studied, the GRU model demonstrates a clear positive correlation between time-series
length and performance. The results indicate that longer input sequences provide richer temporal context, which enhances the GRU’s ability to
capture complex temporal dependencies and improves predictive accuracy consistently across all forecasting horizons.
For the Conv-LSTM model, a similar trend is observed, albeit with diminishing returns as the time-series length increases from 12 to 24. While
both 12 and 24-length time series yield competitive results, the latter shows slight advantages, particularly for longer-term forecasts.
On the other hand, the Conv-GRU model exhibits an unexpected behavior, where mid-range time-series lengths (12) outperform both shorter (6)
and longer (24) inputs. This suggests that the Conv-GRU may balance the trade-off between overfitting to longer sequences and underutilizing
shorter ones. 
The U-TAE model shows competitive performance with both 12 and 24-length time series, although the gap between these configurations widens
for longer forecasting horizons. This behavior implies that while the U-TAE model benefits from longer input sequences for more extended
predictions, it does not fully exploit the additional temporal information as effectively.
FireCastNet demonstrates stability and robustness compared to the other architectures. Regardless of the forecasting horizon, the model
maintains relatively consistent performance, particularly when trained with the longest time-series length of 24. This suggests that
FireCastNet effectively leverages the full temporal context.

\subsection{Time Overlap}

In this section, we conduct an ablation study to investigate the effect of varying sample overlaps between consecutive time series
on model accuracy, as measured by the Area Under the Precision-Recall Curve (AUPRC). The goal is to understand the role of
overlapping temporal context in improving model performance and its implications for training efficiency and predictive accuracy.
In the original setup, for a time series of length $t$, each sample overlaps with $t-1$ identical points from the subsequent sample. 
This ensures that consecutive input sequences share nearly the entire temporal context, differing by only a single time step.
To systematically evaluate the effect of overlap, we consider scenarios where the overlap is reduced. Specifically, for time series
with $t=24$, we analyze configurations where consecutive samples share 3, 6, 12, 18, and including the original full-overlap
configuration of $t-1=23$ samples

The results, illustrated in Figure~\ref{fig:all:ts24:globe:auprc:overlap}, reveal a clear proportional relationship between the degree of
overlap and the resulting AUPRC scores. Higher overlaps consistently yield better AUPRC values, underscoring the importance of preserving
shared temporal information between samples. Using the highest possible overlap ($t-1=23$) offers several key advantages such as enabling
the model to capture subtle temporal patterns and correlations, particularly for dynamic and time-sensitive phenomena and leveraging
information from adjacent time steps. 

While higher overlaps do result in a greater computational burden due to increased redundancy in input samples, the benefits in terms of
predictive accuracy and model robustness outweigh this drawback in many practical scenarios. The results emphasize that preserving shared
temporal context is a crucial factor for achieving superior accuracy, particularly in tasks requiring fine-grained temporal predictions.

\subsection{Variable Importance} \label{sec:xai}

\begin{figure*}[t]
    \centering
    \includegraphics[width=\columnwidth, keepaspectratio]{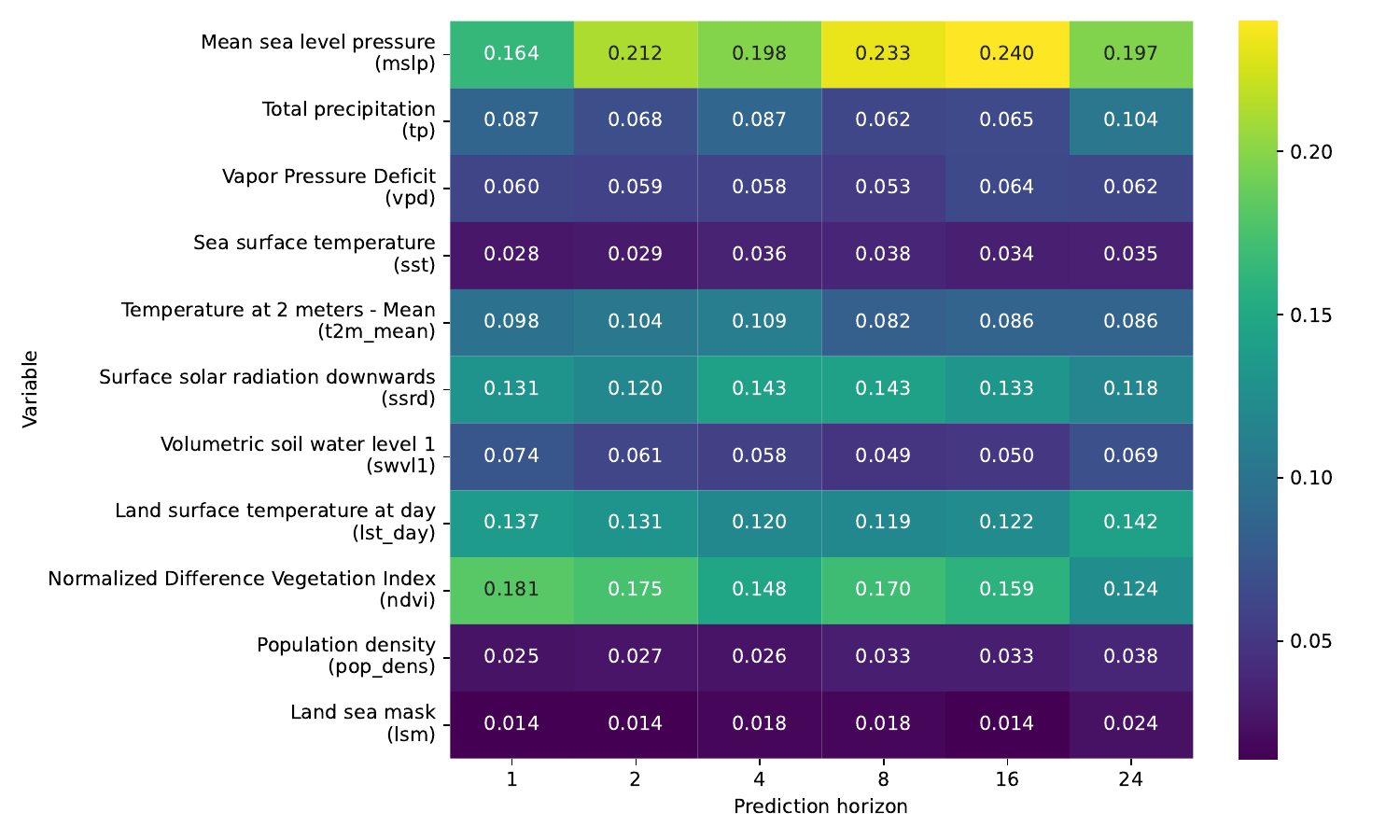} 
    \caption{Feature attributions for fire-danger predictions with the global model across forecasting horizons (1, 2, 4, 8, 16, 24 8-days).
    Each column sums to 1.0, indicating the relative importance of each variable for the respective horizon.
    The results where obtained using Integrated Gradients~\cite{sundararajan2017axiomatic}.
    }
    \label{fig:variables-importances}
\end{figure*}

To understand which input variables are most influential in our FireCastNet predictions, we implemented feature attribution using the
Integrated Gradients~\cite{sundararajan2017axiomatic} technique. For the computation of the gradients, we used zero baseline and performed
200 steps. The resulting feature attributions for the global model are presented in Figure~\ref{fig:variables-importances}.
These attributions indicate the relative importance of each variable in determining the final fire-danger prediction across different
forecasting horizons. Each column in Figure~\ref{fig:variables-importances} corresponds to predictions for a specific horizon and sums
up to 1.0. 

We assessed the importance of 10 input variables and one static variable (Land Sea Mask, lsm)
across six prediction horizons: 1, 2, 4, 8, 16, and 24 8-day periods.
The Mean sea level pressure (mslp) and the Normalized Difference Vegetation Index (ndvi) consistently appear to be the most crucial variables
for predictions across nearly all forecasting horizons. Key variable contributions based on prediction horizon:
\begin{itemize}
    \item Short-Term Forecasts (h=1, 2, 4 8-day periods):
    The highest attributions are consistently found in Normalized Difference Vegetation Index (ndvi) and Mean sea level pressure (mslp). Land surface
    temperature at day (lst\_day) also shows significant influence.
\item Mid-to-Long-Term Forecasts (h=8, 16 8-day periods):
    mslp strongly dominates these horizons, reaching its peak importance at h=16 (0.240). ndvi and Surface solar radiation downwards (ssrd) remain
    key, while variables like Volumetric soil water level 1 (swvl1) show their lowest importance.
\item Longest-Term Forecasts (h=24 8-day periods):
    While mslp remains the single most important variable (0.197), the attribution for ndvi generally decreases. The influence of Land surface
    temperature at day (lst\_day) and Total precipitation (tp) increases significantly compared to mid-range forecasts.
\end{itemize}

Variables with consistently low attribution across all horizons include Sea surface temperature (sst), Population density (pop\_dens), and
the Land sea mask (lsm).
This pattern suggests that while shorter-term predictions rely heavily on immediate environmental conditions (reflected in high initial
NDVI and LST contributions), longer-term seasonal forecasts increasingly depend on broader atmospheric patterns captured by mslp, and the
combined influence of long-term moisture availability (tp) and surface heat (lst\_day).

\subsection{Global Model per GFED Regions}

%
%

%

\newcolumntype{C}[1]{>{\centering\arraybackslash}p{#1}}
\begin{table*}[p]
    \begin{tabular}{|r|C{6em}|c|c|c|c|c|c|}
        \toprule
        \multirow{2}{*}{\textbf{GFED Region}} & \multirow{2}{7em}{\textbf{Fraction of \newline burned areas}} & \multicolumn{6}{c|}{\textbf{Horizon}}\\
        \cline{3-8}
         & & 1 & 2 & 4 & 8 & 16 & 24 \\
        \midrule
        Boreal North America    (BONA) & 0.924\% & 0.04 & 0.03 & 0.03 & 0.03 & 0.02 & 0.02\\
        Temperate North America (TENA) & 1.986\% & 0.26 & 0.27 & 0.26 & 0.26 & 0.25 & 0.26\\
        Central America (CEAM) & 2.137\% & 0.55 & 0.54 & 0.53 & 0.53 & 0.55 & 0.55\\
        Northern Hemisphere South America (NHSA) & 2.673\% & 0.67 & 0.65 & 0.64 & 0.63 & 0.61 & 0.61\\
        Southern Hemisphere South America (SHSA) & 15.619\% & 0.44 & 0.43 & 0.43 & 0.42 & 0.41 & 0.41\\
        Europe (EURO) & 0.857\% & 0.20 & 0.20 & 0.19 & 0.18 & 0.17 & 0.19\\
        Middle East (MIDE) & 1.029\% & 0.29 & 0.30 & 0.29 & 0.30 & 0.29 & 0.26\\
        Northern Hemisphere Africa (NHAF) & 20.749\% & 0.74 & 0.73 & 0.72 & 0.71 & 0.71 & 0.72\\
        Southern Hemisphere Africa (SHAF) & 29.988\% & 0.85 & 0.84 & 0.84 & 0.83 & 0.83 & 0.84\\
        Boreal Asia (BOAS) & 4.072\%  & 0.14 & 0.13 & 0.13 & 0.13 & 0.13 & 0.14\\
        Central Asia (CEAS) & 8.264\% & 0.27 & 0.26 & 0.27 & 0.26 & 0.25 & 0.24\\
        Southeast Asia (SEAS) & 5.764\% & 0.64 & 0.63 & 0.62 & 0.61 & 0.60 & 0.61\\
        Equatorial Asia (EQAS) & 1.089\% & 0.49 & 0.50 & 0.48 & 0.44 & 0.40 & 0.40\\
        Australia and New Zealand (AUST) & 4.849\% & 0.31 & 0.31 & 0.30 & 0.30 & 0.29 & 0.32\\
        \bottomrule
    \end{tabular}
    \centering
    \caption{Global model performance per GFED region for different prediction horizons. The fraction of burned areas
             represents the percentage of wildfires in a specific region relative to the total number of fires globally.}  
    \label{tab:firecastnet:gfed:auprc}
\end{table*}

\begin{figure*}[p]
    \begin{subfigure}[t]{0.24\textwidth}
        \centering
        \begin{tikzpicture}
            \begin{axis}[
                width=\linewidth,
                height=0.75\linewidth,
                title={BONA},
                ymin=0, ymax=1,
                ylabel={AUPRC},
                xtick={1,2,4,8,16,24},
                ytick={0,0.2,0.4,0.6,0.8,1},
                tick label style={font=\tiny},
                label style={font=\tiny},
                title style={font=\footnotesize},
            ]
            \addplot+[mark=*, mark size=1.5pt, color=blue] coordinates {
                (1,0.0364) (2,0.0312) (4,0.0251) (8,0.0254) (16,0.0217) (24,0.0234)
            };
            \addplot[dashed, thick, red] coordinates {(1,0.0075) (24,0.0075)};
            \addplot[dashed, thick, green!60!black] coordinates {(1,0.0001) (24,0.0001)};
            \end{axis}
        \end{tikzpicture}
    \end{subfigure}%
    \hfill
    \begin{subfigure}[t]{0.24\textwidth}
        \centering
        \begin{tikzpicture}
            \begin{axis}[
                width=\linewidth,
                height=0.75\linewidth,
                title={TENA},
                ymin=0, ymax=1,
                xtick={1,2,4,8,16,24},
                ytick={0,0.2,0.4,0.6,0.8,1},
                tick label style={font=\tiny},
                label style={font=\tiny},
                title style={font=\footnotesize},
            ]
            \addplot+[mark=*, mark size=1.5pt, color=blue] coordinates {
                (1,0.2608) (2,0.2673) (4,0.2621) (8,0.2571) (16,0.2539) (24,0.2573)
            };
            \addplot[dashed, thick, red] coordinates {(1,0.0442) (24,0.0442)};
            \addplot[dashed, thick, green!60!black] coordinates {(1,0.0842) (24,0.0842)};
            \end{axis}
        \end{tikzpicture}
    \end{subfigure}%
    \hfill
    \begin{subfigure}[t]{0.24\textwidth}
        \centering
        \begin{tikzpicture}
            \begin{axis}[
                width=\linewidth,
                height=0.75\linewidth,
                title={CEAM},
                ymin=0, ymax=1,
                xtick={1,2,4,8,16,24},
                ytick={0,0.2,0.4,0.6,0.8,1},
                tick label style={font=\tiny},
                label style={font=\tiny},
                title style={font=\footnotesize},
            ]
            \addplot+[mark=*, mark size=1.5pt, color=blue] coordinates {
                (1,0.5499) (2,0.5379) (4,0.5318) (8,0.5328) (16,0.5456) (24,0.5490)
            };
            \addplot[dashed, thick, red] coordinates {(1,0.1854) (24,0.1854)};
            \addplot[dashed, thick, green!60!black] coordinates {(1,0.1996) (24,0.1996)};
            \end{axis}
        \end{tikzpicture}
    \end{subfigure}%
    \hfill
    \begin{subfigure}[t]{0.24\textwidth}
        \centering
        \begin{tikzpicture}
            \begin{axis}[
                width=\linewidth,
                height=0.75\linewidth,
                title={NHSA},
                ymin=0, ymax=1,
                xtick={1,2,4,8,16,24},
                ytick={0,0.2,0.4,0.6,0.8,1},
                tick label style={font=\tiny},
                label style={font=\tiny},
                title style={font=\footnotesize},
            ]
            \addplot+[mark=*, mark size=1.5pt, color=blue] coordinates {
                (1,0.6643) (2,0.6475) (4,0.6360) (8,0.6288) (16,0.6067) (24,0.6047)
            };
            \addplot[dashed, thick, red] coordinates {(1,0.2575) (24,0.2575)};
            \addplot[dashed, thick, green!60!black] coordinates {(1,0.2560) (24,0.2560)};
            \end{axis}
        \end{tikzpicture}
    \end{subfigure}%
    \vfill
    \begin{subfigure}[t]{0.24\textwidth}
    \centering
    \begin{tikzpicture}
    \begin{axis}[width=\linewidth, height=0.75\linewidth, title={SHSA}, ymin=0, ymax=1, 
        ylabel={AUPRC}, xtick={1,2,4,8,16,24}, ytick={0,0.2,...,1}, tick label style={font=\tiny}, label style={font=\tiny}, title style={font=\footnotesize}]
    \addplot+[mark=*, mark size=1.5pt, color=blue] coordinates {(1,0.4382)(2,0.4321)(4,0.4285)(8,0.4184)(16,0.4119)(24,0.4083)};
    \addplot[dashed, thick, red] coordinates {(1,0.1527)(24,0.1527)};
    \addplot[dashed, thick, green!60!black] coordinates {(1,0.1459)(24,0.1459)};
    \end{axis}
    \end{tikzpicture}
    \end{subfigure}%
    \hfill
    \begin{subfigure}[t]{0.24\textwidth}
    \centering
    \begin{tikzpicture}
    \begin{axis}[width=\linewidth, height=0.75\linewidth, title={EURO}, ymin=0, ymax=1, 
        xtick={1,2,4,8,16,24}, ytick={0,0.2,...,1}, tick label style={font=\tiny}, label style={font=\tiny}, title style={font=\footnotesize}]
    \addplot+[mark=*, mark size=1.5pt, color=blue] coordinates {(1,0.1959)(2,0.1946)(4,0.1860)(8,0.1832)(16,0.1737)(24,0.1864)};
    \addplot[dashed, thick, red] coordinates {(1,0.0489)(24,0.0489)};
    \addplot[dashed, thick, green!60!black] coordinates {(1,0.0269)(24,0.0269)};
    \end{axis}
    \end{tikzpicture}
    \end{subfigure}%
    \hfill
    \begin{subfigure}[t]{0.24\textwidth}
    \centering
    \begin{tikzpicture}
    \begin{axis}[width=\linewidth, height=0.75\linewidth, title={MIDE}, ymin=0, ymax=1, 
        xtick={1,2,4,8,16,24}, ytick={0,0.2,...,1}, tick label style={font=\tiny}, label style={font=\tiny}, title style={font=\footnotesize}]
    \addplot+[mark=*, mark size=1.5pt, color=blue] coordinates {(1,0.2846)(2,0.2957)(4,0.2855)(8,0.2945)(16,0.2870)(24,0.2574)};
    \addplot[dashed, thick, red] coordinates {(1,0.0776)(24,0.0776)};
    \addplot[dashed, thick, green!60!black] coordinates {(1,0.0334)(24,0.0334)};
    \end{axis}
    \end{tikzpicture}
    \end{subfigure}%
    \hfill
    \begin{subfigure}[t]{0.24\textwidth}
    \centering
    \begin{tikzpicture}
    \begin{axis}[width=\linewidth, height=0.75\linewidth, title={NHAF}, ymin=0, ymax=1, 
        xtick={1,2,4,8,16,24}, ytick={0,0.2,...,1}, tick label style={font=\tiny}, label style={font=\tiny}, title style={font=\footnotesize}]
    \addplot+[mark=*, mark size=1.5pt, color=blue] coordinates {(1,0.7371)(2,0.7279)(4,0.7163)(8,0.7097)(16,0.7140)(24,0.7181)};
    \addplot[dashed, thick, red] coordinates {(1,0.3094)(24,0.3094)};
    \addplot[dashed, thick, green!60!black] coordinates {(1,0.4437)(24,0.4437)};
    \end{axis}
    \end{tikzpicture}
    \end{subfigure}
    \vfill
    \begin{subfigure}[t]{0.24\textwidth}
    \centering
    \begin{tikzpicture}
    \begin{axis}[width=\linewidth, height=0.75\linewidth, title={SHAF}, ymin=0, ymax=1, 
        xlabel={\phantom{Horizon}}, 
        ylabel={AUPRC},
    xtick={1,2,4,8,16,24}, ytick={0,0.2,...,1}, tick label style={font=\tiny}, label style={font=\tiny}, title style={font=\footnotesize}]
    \addplot+[mark=*, mark size=1.5pt, color=blue] coordinates {(1,0.8465)(2,0.8432)(4,0.8409)(8,0.8303)(16,0.8308)(24,0.8371)};
    \addplot[dashed, thick, red] coordinates {(1,0.3733)(24,0.3733)};
    \addplot[dashed, thick, green!60!black] coordinates {(1,0.5639)(24,0.5639)};
    \end{axis}
    \end{tikzpicture}
    \end{subfigure}%
    \hfill
    \begin{subfigure}[t]{0.24\textwidth}
    \centering
    \begin{tikzpicture}
    \begin{axis}[width=\linewidth, height=0.75\linewidth, title={BOAS}, ymin=0, ymax=1, 
        xlabel={Horizon}, 
        xtick={1,2,4,8,16,24}, ytick={0,0.2,...,1}, tick label style={font=\tiny}, label style={font=\tiny}, title style={font=\footnotesize}]
    \addplot+[mark=*, mark size=1.5pt, color=blue] coordinates {(1,0.1420)(2,0.1336)(4,0.1264)(8,0.1313)(16,0.1319)(24,0.1400)};
    \addplot[dashed, thick, red] coordinates {(1,0.0331)(24,0.0331)};
    \addplot[dashed, thick, green!60!black] coordinates {(1,0.0161)(24,0.0161)};
    \end{axis}
    \end{tikzpicture}
    \end{subfigure}%
    \hfill
    \begin{subfigure}[t]{0.24\textwidth}
    \centering
    \begin{tikzpicture}
    \begin{axis}[width=\linewidth, height=0.75\linewidth, title={CEAS}, ymin=0, ymax=1, 
        xlabel={Horizon}, 
        xtick={1,2,4,8,16,24}, ytick={0,0.2,...,1}, tick label style={font=\tiny}, label style={font=\tiny}, title style={font=\footnotesize}]
    \addplot+[mark=*, mark size=1.5pt, color=blue] coordinates {(1,0.2736)(2,0.2594)(4,0.2647)(8,0.2565)(16,0.2466)(24,0.2399)};
    \addplot[dashed, thick, red] coordinates {(1,0.0672)(24,0.0672)};
    \addplot[dashed, thick, green!60!black] coordinates {(1,0.0602)(24,0.0602)};
    \end{axis}
    \end{tikzpicture}
    \end{subfigure}%
    \hfill
    \begin{subfigure}[t]{0.24\textwidth}
    \centering
    \begin{tikzpicture}
    \begin{axis}[width=\linewidth, height=0.75\linewidth, title={SEAS}, ymin=0, ymax=1, 
        xlabel={\phantom{Horizon}}, 
        xtick={1,2,4,8,16,24}, ytick={0,0.2,...,1}, tick label style={font=\tiny}, label style={font=\tiny}, title style={font=\footnotesize}]
    \addplot+[mark=*, mark size=1.5pt, color=blue] coordinates {(1,0.6410)(2,0.6312)(4,0.6206)(8,0.6087)(16,0.5988)(24,0.6069)};
    \addplot[dashed, thick, red] coordinates {(1,0.2500)(24,0.2500)};
    \addplot[dashed, thick, green!60!black] coordinates {(1,0.2964)(24,0.2964)};
    \end{axis}
    \end{tikzpicture}
    \end{subfigure}
    \vfill
    \begin{subfigure}[t]{0.24\textwidth}
    \centering
    \begin{tikzpicture}
    \begin{axis}[width=\linewidth, height=0.75\linewidth, title={EQAS}, ymin=0, ymax=1, xlabel={Horizon}, ylabel={AUPRC},
    xtick={1,2,4,8,16,24}, ytick={0,0.2,...,1}, tick label style={font=\tiny}, label style={font=\tiny}, title style={font=\footnotesize}]
    \addplot+[mark=*, mark size=1.5pt, color=blue] coordinates {(1,0.4874)(2,0.4953)(4,0.4784)(8,0.4360)(16,0.4044)(24,0.4022)};
    \addplot[dashed, thick, red] coordinates {(1,0.1360)(24,0.1360)};
    \addplot[dashed, thick, green!60!black] coordinates {(1,0.0963)(24,0.0963)};
    \end{axis}
    \end{tikzpicture}
    \end{subfigure}%
    \hfill
    \begin{subfigure}[t]{0.24\textwidth}
    \centering
    \begin{tikzpicture}
    \begin{axis}[
        width=\linewidth, 
        height=0.75\linewidth, 
        title={AUST}, 
        ymin=0, 
        ymax=1,
        xlabel={Horizon}, 
        xtick={1,2,4,8,16,24}, 
        ytick={0,0.2,...,1}, 
        tick label style={font=\tiny}, 
        label style={font=\tiny}, 
        title style={font=\footnotesize}
    ]
    \addplot+[mark=*, mark size=1.5pt, color=blue] coordinates {(1,0.3095)(2,0.3046)(4,0.3036)(8,0.3015)(16,0.2919)(24,0.3180)};
    \addplot[dashed, thick, red] coordinates {(1,0.0577)(24,0.0577)};
    \addplot[dashed, thick, green!60!black] coordinates {(1,0.0787)(24,0.0787)};
    \end{axis}
    \end{tikzpicture}
    \end{subfigure}

    \caption{Global model AUPRC performance across horizons per GFED region. 
    Dashed lines represent the any-fire (red) and majority (green) baselines.} \label{fig:firecastnet:gfed:subplots-simple}
\end{figure*}

In this section, we present predictions derived using regional masks from the Global Fire Emissions
Database (GFED), as provided by the SeasFire Datacube~\cite{karasante2023seasfire}. These regional masks allow for a more
granular analysis of fire activity by focusing on specific geographic areas. To produce region-specific predictions,
we compute the Area Under the Precision-Recall Curve (AUPRC) for each region by applying the GFED masks to both the
model's output and the ground truth burned area data.

The motivation for evaluating the model's performance across different GFED regions stems from the fact that 
the performance of the global model, as can be seen in Table~\ref{tab:all:ts24:globe:auprc}, does not always drop significantly
as the prediction horizon increases. Thus, we hypothesize that the model's performance may vary across different regions depending 
on the number of samples in our dataset. This is particularly relevant for regions with distinct fire regimes, 
where the model's ability to generalize may be influenced by the availability of training data and the specific characteristics
of each region's fire patterns. Notable region is Africa, which has a significant number of wildfires. 

Table~\ref{tab:firecastnet:gfed:auprc} and Figure~\ref{fig:firecastnet:gfed:subplots-simple} summarize the AUPRC results
across all GFED regions and for multiple time horizons, providing a comprehensive overview of the model's performance in
diverse settings. In the same table we have also added the fractions of burned areas per GFED region. These represent
the percentages of wildfires in a specific region relative to the total number of fires globally, calculated from
2002-01-01 to 2020-01-01. Notably, FireCastNet demonstrates superior accuracy in burned area forecasts, particularly
in Southeast Asia, Northern Hemisphere South America, and Africa, highlighting its effectiveness in regions with
distinct fire regimes and challenges.

On the other hand, some regions such as Northern Hemisphere Africa (NHAF) and Southern Hemisphere Africa (SHAF) exhibit 
different behavior than regions such as Northern Hemisphere South America (NHSA) and Southern Hemisphere South America (SHSA).
NHAF and SHAF show a drop in performance as the prediction horizon increases, while NHSA and SHSA maintain
relatively stable performance across horizons. This suggests that in the case of Africa the model seems to learn some 
advanced seasonality pattern instead of learning how to predict the burned areas.
This is further supported by the fact that the model's performance in Africa is significantly better than in
other regions, such as Southeast Asia (SEAS) and Equatorial Asia (EQAS), where the model's performance drops significantly
as the prediction horizon increases. This discussion motivates Section~\ref{sec:lam}.

\subsection{Local Area Modelling} \label{sec:lam}

The Local Area Modelling (LAM) approach~\cite{oskarsson2023graph, lam2023learning} extends the uniform global strategy
by implementing region-specific
training. Building upon the hierarchical multi-resolution mesh construction, where each target GFED region receives
fine-grained mesh granularity while surrounding areas transition to progressively coarser levels, we train dedicated
FireCastNet models for individual regions of interest. This localized training strategy offers several key advantages over
the global model approach. 

First, by constraining the training domain to a specific region, the model can better capture the unique fire-climate
relationships and local environmental drivers that characterize different biomes and geographical areas. Second, the
computational focus on smaller geographical domains allows for more efficient use of the model's capacity.
Third, the LAM approach enables enhanced spatial resolution within the region of interest while maintaining computational
tractability. The adaptive mesh strategy concentrates the finest mesh granularity within the target region, providing
higher spatial fidelity for local fire prediction while the coarser surrounding mesh levels capture the necessary global
teleconnections and boundary conditions.

\begin{figure*}[t]
    \centering
    \includegraphics[width=0.6\columnwidth, keepaspectratio]{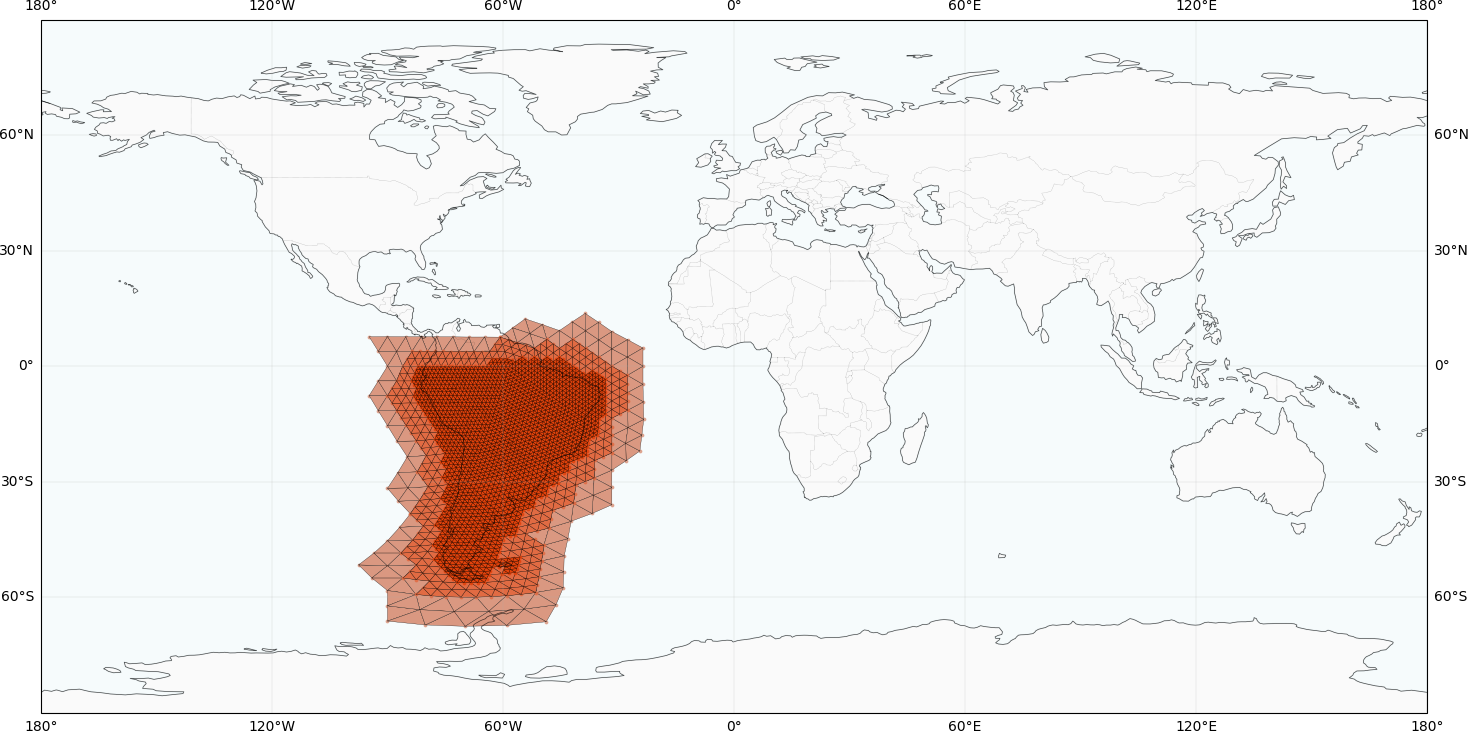} 
    \hspace{0.05\columnwidth}
    \includegraphics[width=0.30\columnwidth, keepaspectratio]{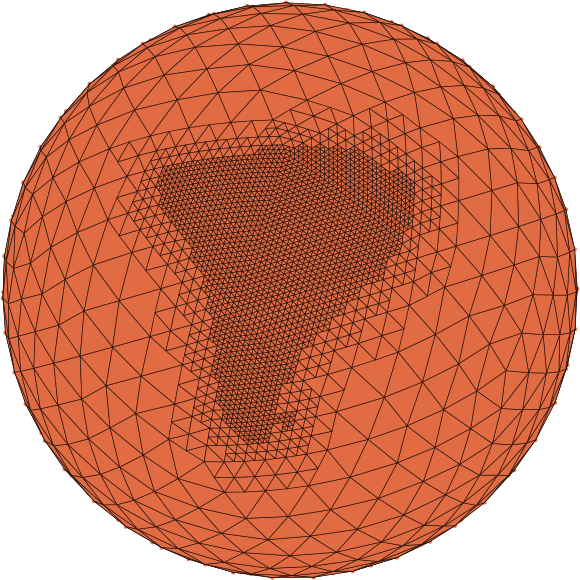} 
    \caption{Example of mesh using the Local Area Modelling (LAM) approach for the SHSA GFED Region (Southern Hemisphere South America).
    The left image shows the flat mesh representation around the target region. While not shown the grid expands to the entire globe
    as can be seen in the right image, which shows the spherical mesh representation.    
    The target region is represented at the sixth level of detail, while the surrounding areas are represented
    at the third level of detail. The mesh is adapted to focus on the target region, allowing for higher resolution
    predictions in that area while maintaining a coarser resolution in the surrounding regions.
    The transition from finer to coarser mesh levels is gradual, using buffer regions of size 400-800km, ensuring a smooth transition and
    preserving the spatial coherence of the model's predictions.}
    \label{fig:lam:mesh:example}
\end{figure*}

Figure~\ref{fig:lam:mesh:example} illustrates the adaptive mesh strategy employed in the LAM approach, 
highlighting the varying levels of granularity across different regions. The figure demonstrates how
the mesh is refined within the target region while coarser resolutions are used in surrounding areas,
effectively balancing detail and computational efficiency. We reach the sixth level of detail in the
target region, while the surrounding areas are represented at the third level of detail (levels here correspond to 
the different mesh resolutions in the original GraphCast mesh construction). The change from 
finer to coarse mesh levels is gradual, ensuring a smooth transition and preserving the spatial coherence of the model's
predictions.

\newcolumntype{C}[1]{>{\centering\arraybackslash}p{#1}}
\begin{table*}[t]
    \begin{tabular}{|r|C{6em}|c|c|c|c|c|c|}
        \toprule
        \multirow{2}{*}{\textbf{GFED Region}} & \multirow{2}{7em}{\textbf{Fraction of \newline burned areas}} & \multicolumn{6}{c|}{\textbf{Horizon}}\\
        \cline{3-8}
         & & 1 & 2 & 4 & 8 & 16 & 24 \\
        \midrule
        Boreal North America    (BONA) & 0.924\% & 0.04 & 0.04 & 0.03 & 0.02  & 0.03  & 0.03 \\
        Temperate North America (TENA) & 1.986\% & 0.28 & 0.27  & 0.27  & 0.28  & 0.29  & 0.29 \\
        Central America (CEAM) & 2.137\% & 0.40 & 0.39 & 0.37  & 0.38  & 0.39  & 0.38 \\
        Northern Hemisphere South America (NHSA) & 2.673\% & 0.56  & 0.55  & 0.55 & 0.54 & 0.54  & 0.54 \\
        Southern Hemisphere South America (SHSA) & 15.619\% & 0.47 & 0.47  & 0.47  & 0.46  & 0.45  & 0.44 \\
        Europe (EURO) & 0.857\% & 0.24 & 0.24 & 0.24 & 0.21 & 0.24  & 0.20 \\
        Middle East (MIDE) & 1.029\% & 0.31  & 0.31 & 0.29  & 0.33 & 0.29 & 0.28 \\
        Northern Hemisphere Africa (NHAF) & 20.749\% & 0.78  & 0.77 & 0.75 & 0.74 & 0.75  & 0.75 \\
        Southern Hemisphere Africa (SHAF) & 29.988\% & 0.86  & 0.86 & 0.86 & 0.85  & 0.85 & 0.85 \\
        Boreal Asia (BOAS) & 4.072\%  & 0.09  & 0.10 & 0.07  & 0.06 & 0.06  & 0.07 \\
        Central Asia (CEAS) & 8.264\% & 0.31  & 0.31  & 0.31  & 0.30 & 0.28  & 0.29 \\
        Southeast Asia (SEAS) & 5.764\% & 0.57  & 0.57  & 0.55  & 0.55  & 0.52 & 0.54 \\
        Equatorial Asia (EQAS) & 1.089\% & 0.50  & 0.49 & 0.48  & 0.46 & 0.47 & 0.48 \\
        Australia and New Zealand (AUST) & 4.849\% & 0.25  & 0.26 & 0.24  & 0.24 & 0.25 & 0.26 \\
        \bottomrule
    \end{tabular}
    \centering
    \caption{Local Area Model performance per GFED region for different prediction horizons. A different
    model is trained for each region, using the same training data as the global model, but
    focusing on the region of interest.
     }      
    \label{tab:lam:gfed:auprc}
\end{table*}

Training these localized models involves adapting the FireCastNet loss function to compute the loss only 
on the region of interest. Table~\ref{tab:lam:gfed:auprc} presents the results. 
Again each table entry corresponds to a separate model trained on the region of interest for that particular 
prediction horizon, using the same training data as the global model. 
The results can be compared with the global model results presented in Table~\ref{tab:firecastnet:gfed:auprc}, 
indicating a varied impact on
performance across different GFED regions.
Several regions exhibit a notable improvement in AUPRC scores with the LAM models. For instance, regions in Africa
(NHAF and SHAF), Europe (EURO), the Middle East (MIDE), and parts of the Americas (TENA, SHSA) and Asia (EQAS, CEAS) show
enhanced predictive performance. Conversely, for some regions, the global model outperforms the specialized LAM models.
This is observed in regions such as Boreal North America (BONA), Central America (CEAM), Northern Hemisphere South
America (NHSA), Boreal Asia (BOAS), Southeast Asia (SEAS), and Australia (AUST).

In these cases, it is plausible that long-range teleconnections, which are better captured by the global model, play a
more dominant role in driving fire activity than local factors. The global model's ability to learn from a wider range of
fire regimes and climatic conditions across the entire planet may provide a more robust and generalizable prediction
for these specific areas. This highlights a key trade-off between local specialization and the comprehensive global context.

Examining the relationship between LAM performance improvements and the fraction of burned areas in each region
(which serves as a proxy for data availability), we find no clear correlation between the number of samples and
whether LAM models outperform the global model. While some high-activity regions like Africa (NHAF, SHAF) benefit
from LAM, other regions with substantial fire activity such as Southeast Asia (SEAS) and Australia (AUST) do not
show improvement. Similarly, regions with limited fire activity like Europe (EURO) and Temperate North America
(TENA) still benefit from the LAM approach. This lack of observable correlation suggests that the effectiveness
of LAM models depends on factors beyond simple data availability, likely related to the underlying fire dynamics
and their relationship to local versus global climate drivers in each region.

\section{Conclusion}

The findings of this study underscore the potential of machine learning models in advancing seasonal wildfire
forecasting. Our architecture, FireCastNet, integrates 3D convolutional encoding with GraphCast, in order to
capture the spatio-temporal context of wildfires, enabling predictive analysis across various forecasting
horizons. Our architecture is clearly more beneficial than both baseline and contemporary methods for
spatio-temporal predictions and burned area prediction via semantic segmentation.

One significant insight derived from our investigation is the importance of temporal context. The results
clearly show that longer input time-series enhance predictive robustness across forecasting horizons. This
finding aligns with the intuition that wildfires are influenced by accumulated climatic, vegetative, and
human-related factors, which can be utilized more effectively by adding a deeper temporal history.

This outcome shows the benefits of architectures that can process both local and regional patterns,
highlighting the interconnected nature of wildfire phenomena where adjacent regions' conditions can
influence fire behavior. 
    
    \section*{Acknowledgment}

    This work has received funding from the SeasFire project, which focuses on Earth System Deep Learning for
    Seasonal Fire Forecasting and is funded by the European Space Agency (ESA) under the ESA Future EO-1 Science
    for Society Call.

\end{document}